\documentclass[conference,a4paper]{IEEEtran}
\IEEEoverridecommandlockouts

\usepackage[hidelinks]{hyperref}
\usepackage[cmex10]{amsmath}
\usepackage{amssymb,amsfonts}
\usepackage{dblfloatfix}

\usepackage[ruled,vlined]{algorithm2e}
\usepackage[usenames,dvipsnames]{color}
\usepackage{graphicx}
\usepackage[font=footnotesize,labelfont=bf]{caption}
\graphicspath{{Figures/PDF/}{Figures/PNG/}}

\usepackage{booktabs}
\usepackage{multirow}
\usepackage{adjustbox}
\usepackage{siunitx}
\usepackage[numbers,compress]{natbib}
\usepackage{texnames}
\usepackage{bm,bbm}
\usepackage{orcidlink}

\begin{document}

\title{MMLGNet: Cross-Modal Alignment of Remote Sensing Data using CLIP}

\author{\IEEEauthorblockN{Aditya Chaudhary \orcidlink{0009-0008-2551-7619}}
	\IEEEauthorblockA{\textit{Dept. of CSE}\\
		The LNMIIT Jaipur\\
        }
        \and
        \IEEEauthorblockN{Sneha Barman \orcidlink{0009-0001-4711-5695}}
	\IEEEauthorblockA{\textit{DAGP}\\
		IIT Dhanbad\\
        }
        \and
        \IEEEauthorblockN{Mainak Singha \orcidlink{0000-0002-7615-2575}}
	\IEEEauthorblockA{\textit{CSRE}\\
		IIT Bombay\\
        }
        \and
        \IEEEauthorblockN{Ankit Jha \orcidlink{0000-0002-1063-8978}}
	\IEEEauthorblockA{\textit{Dept. of CSE}\\
		The LNMIIT Jaipur\\
        }
        \and 
        \IEEEauthorblockN{Girish Mishra \orcidlink{0000-0002-7489-2721}}
	\IEEEauthorblockA{\textit{SAG}\\
		DRDO, Delhi\\
		}
        \and
        \IEEEauthorblockN{Biplab Banerjee \orcidlink{0000-0001-8371-8138}}
	\IEEEauthorblockA{\textit{CSRE}\\
		IIT Bombay\\
        }
}

\maketitle
\makeatletter
\renewenvironment{abstract}{%
  \small 
  \begin{center}%
    {\bfseries \abstractname\vspace{-1ex}\vspace{0pt}}%
  \end{center}%
  \quotation
}{\endquotation}
\makeatother
\begin{abstract}
In this paper, we propose a novel multimodal framework, Multimodal Language-Guided Network (MMLGNet), to align heterogeneous remote sensing modalities like Hyperspectral Imaging (HSI) and LiDAR with natural language semantics using vision-language models such as CLIP. With the increasing availability of multimodal Earth observation data, there is a growing need for methods that effectively fuse spectral, spatial, and geometric information while enabling semantic-level understanding. MMLGNet employs modality-specific encoders and aligns visual features with handcrafted textual embeddings in a shared latent space via bi-directional contrastive learning. Inspired by CLIP’s training paradigm, our approach bridges the gap between high-dimensional remote sensing data and language-guided interpretation. Notably, MMLGNet achieves strong performance with simple CNN-based encoders, outperforming several established multimodal visual-only methods on two benchmark datasets, demonstrating the significant benefit of language supervision. Codes are available at \url{https://github.com/AdityaChaudhary2913/CLIP_HSI}.
\end{abstract}


\section{Introduction}

Remote sensing systems increasingly rely on a diverse set of sensors such as Hyperspectral Imaging (HSI), Multispectral Imaging (MSI), and Light Detection and Ranging (LiDAR), each offering unique advantages in characterizing the Earth's surface. HSI provides fine spectral granularity, enabling material discrimination \cite{hsi}; MSI offers broader spectral coverage with fewer channels and higher spatial resolution; and LiDAR contributes structural and topographic information \cite{lidar}. The fusion of these modalities has shown promise in tasks like land use mapping \cite{koetz2008multi}, crop type classification \cite{kussul2017deep}, and environmental monitoring \cite{yuan2020deep, shu2025deep}. Despite these benefits, combining such heterogeneous data sources presents non-trivial challenges, primarily due to differences in spatial resolution, modality-specific noise, and limited availability of aligned datasets. Traditional fusion approaches such as pixel-level stacking \cite{healey2018mapping}, feature concatenation \cite{chaib2017deep, Jha_2023_WACV}, or decision-level \cite{shen2018comparison} voting suffer from scalability issues and often demand extensive labeled data for supervised learning.

Recent advances in vision-language pretraining, particularly Contrastive Language Image Pretraining (CLIP) \cite{Clip}, have opened a new direction: using textual supervision as a unifying modality to encode visual content semantically. CLIP aligns images and text in a shared embedding space via contrastive learning, allowing zero-shot recognition based on class names or textual prompts. Although widely successful on natural image datasets like ImageNet \cite{imagenet}, CLIP and its adaptations have yet to be deeply explored in the context of multimodal remote sensing data. Several efforts have extended CLIP-like architectures to overhead imagery focusing primarily on RGB data \cite{Singha_2023_CVPR}. However, such methods generally overlook richer modalities like HSI and LiDAR, which offer deeper spectral or structural cues critical for fine-grained semantic understanding. Work like GeoCLIP \cite{geoclip} attempted to incorporate geospatial knowledge into language-image models, but their focus remained on natural color or pseudo-color imagery. Additionally, while a few studies have explored integrating HSI with deep learning for classification, and even fewer have considered language supervision, there is a clear research gap in unifying diverse remote sensing modalities under a vision-language framework.

\begin{figure*}[ht!]
  \centering
  \includegraphics[width=0.85\linewidth]{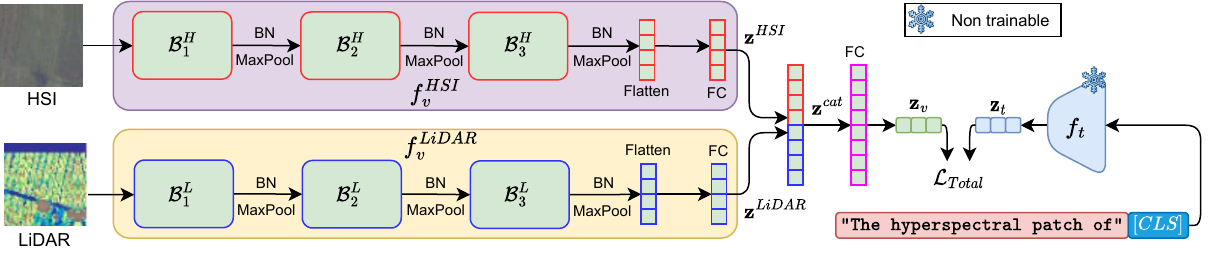}
  \vspace{-0.3cm}
  \caption{Architecture overivew of our multimodal CLIP-based MMLGNet.}
  
  \label{fig:CLIP_MultiModal}
\end{figure*}
In this paper, we propose MMLGNet a simple framework for aligning multimodal remote sensing data with CLIP. It leverages contrastive pretraining to bridge HSI, and LiDAR data with natural language descriptions. The core objective is to align visual features from different modalities with textual embeddings, enabling the learning of robust representations in a shared latent space. We design modality-specific encoders that preserve spectral, spatial, and structural information, followed by a contrastive alignment mechanism adapted from CLIP to guide multimodal visual features toward semantically meaningful language embeddings. This work contributes to the advancement of multimodal foundation models for remote sensing and offers a scalable way to integrate language understanding into geospatial analysis. We summarize our contribution as:\\
\noindent - We propose MMLGNet, a simple and effective framework that aligns heterogeneous remote sensing modalities like HSI and LiDAR along with the handcrafted textual descriptions using contrastive training, enabling semantically enriched feature representations in a shared latent space. \\
\noindent - We design the CNN-based encoders for HSI and LiDAR modalities, demonstrating that language supervision can effectively enhance visual representations without relying on complex architectures. Also, We perform visual-text contrastive alignment using prompts generated for land-cover classes, bridging the semantic gap between raw visual features and natural language semantics.\\
\noindent - We evaluated our proposed MMLGNet on the MUUFL Gulfport and Trento datasets, where it consistently outperforms several established methods that rely solely on visual modalities.

\section{Methodology}

\subsection{Preliminaries}
Let us consider a multimodal RS dataset $\mathcal{D} = \{\mathbf{X}_{\text{HSI}}, \mathbf{X}_{\text{LiDAR}}, \mathbf{Y}\}$, where $\mathbf{X}_{\text{HSI}} \in \mathbb{R}^{H \times W \times B}$ represents the HSI with $B$ spectral bands, $\mathbf{X}_{\text{LiDAR}} \in \mathbb{R}^{H \times W \times 1}$ denotes the corresponding registered LiDAR-derived elevation map, and $\mathbf{Y} \in \{1, 2, \dots, C\}^{H \times W}$ is the ground-truth label map with $C$ semantic classes. Here, $H$ and $W$ denote the height and width of the spatial domain. To perform pixel-level classification, non-overlapping patches of size $P \times P$ are extracted from both modalities around each pixel, resulting in paired inputs: $\mathbf{p}^{\text{HSI}} \in \mathbb{R}^{B \times P \times P}$ and $\mathbf{p}^{\text{LiDAR}} \in \mathbb{R}^{1 \times P \times P}$. The goal is to learn a mapping function $\varphi_{\theta}: (\mathbf{p}^{\text{HSI}}, \mathbf{p}^{\text{LiDAR}}) \mapsto y$ that predicts the semantic label $y$ for each pixel using the complementary spectral and structural information, with additional guidance from a language-based supervision framework.

\subsection{Our proposed MMLGNet}
Our proposed Multimodal Language-Guided Network (MMLGNet) is designed to learn a joint embedding space where semantically similar visual data from HSI and LiDAR sensors are aligned with natural language descriptions, shown in Figure~\ref{fig:CLIP_MultiModal}. It comprises of modality-specific visual encoders parameterized by $\theta_v^{H}$ for HSI encoding and $\theta_v^{L}$ for LiDAR encoding, a fusion module, and a pretrained frozen text encoder $f_t$. The objective is to learn the parameters $\theta_v = \{\theta_v^{H}, \theta_v^{L}\}$ that maps the paired visual patches $(x_\mathbf{p}^{\text{HSI}}, x_\mathbf{p}^{\text{LiDAR}})$ to a visual embedding ${\mathbf{z}}_v$ that is close to the text embedding ${\mathbf{z}}_t$ of the corresponding ground-truth class description.

\subsubsection{Modality-Specific Visual Feature Extraction}
To process the heterogeneous inputs, we employ two specialized Convolutional Neural Networks (CNNs), an HSI encoder $E_H$ and a LiDAR encoder $E_L$, which constitute our visual encoder backbone with parameters $\theta_v \in \{\theta_v^{H}, \theta_v^{L}\}$. Let a single convolutional block, $\mathcal{B}$, be defined as the composition of 2D convolution ($\ast$), Batch Normalization (BN), a non-linear activation function ($\sigma$), and Max Pooling. For an input feature map $\mathbf{X}$ and learnable weights $\mathbf{W}$ and bias $\mathbf{b}$, the operation is:
\begin{equation}
    \mathcal{B}(\mathbf{X}; \mathbf{W}, \mathbf{b}) = \text{MaxPool}(\sigma(\text{BN}(\mathbf{X} \ast \mathbf{W} + \mathbf{b})))
\end{equation}
Given an HSI patch $x_\mathbf{p}^{\text{HSI}} \in \mathbb{R}^{B \times p \times p}$, the HSI encoder is a composition of three such convolutional blocks ($\mathcal{B}_1^H, \mathcal{B}_2^H, \mathcal{B}_3^H$), followed by a flatten operation and a fully connected (FC) layer i.e. linear layer. This process extracts a dense feature embedding, denoted in Equation \ref{eq:fhsi}, where the parameters $\theta_H$ consist of the weights and biases of the CNN and FC layers.
\begin{equation}\label{eq:fhsi}
    \mathbf{z}^{\text{HSI}} = \text{FC}(\text{Flatten}(\mathcal{B}_3^H(\mathcal{B}_2^H(\mathcal{B}_1^H(\mathbf{p}^{\text{HSI}}))))) \in \mathbb{R}^{D/2}
\end{equation}

Similarly, for a LiDAR patch $\mathbf{p}^{\text{LiDAR}} \in \mathbb{R}^{1 \times P \times P}$, the LiDAR encoder $E_L$ applies an analogous sequence of operations with its own set of parameters $\theta_L$ to produce a structural feature embedding defined in Equation \ref{eq:flid}.
\begin{equation}\label{eq:flid}
    \mathbf{z}^{\text{LiDAR}} = \text{FC}(\text{Flatten}(\mathcal{B}_3^L(\mathcal{B}_2^L(\mathcal{B}_1^L(\mathbf{p}^{\text{LiDAR}}))))) \in \mathbb{R}^{D/2}
\end{equation}

\subsubsection{Multimodal Visual Feature Fusion}
The core of our network is an explicit fusion module that learns a joint representation from the individual modality features. The HSI and LiDAR feature embeddings are first concatenated to form a unified multimodal vector $\mathbf{z}^{\text{cat}} = \mathbf{z}^{\text{HSI}} \oplus \mathbf{z}^{\text{LiDAR}} \in \mathbb{R}^{D}$,
where $\oplus$ denotes the concatenation operation. The fused multimodal visual feature vector is then transformed by a linear fusion layer, parameterized by a weight matrix $\mathbf{w}_{\text{fuse}} \in \mathbb{R}^{D \times D}$ and bias $\mathbf{b}_{\text{fuse}} \in \mathbb{R}^{D}$. This layer projects the concatenated features into the final shared embedding space, producing the visual embedding $\mathbf{z}_v = \mathbf{w}_{\text{fuse}} \cdot \mathbf{z}^{\text{cat}} + \mathbf{b}_{\text{fuse}}$.

\subsubsection{Language-Vision Alignment}
We proposed to align the learnt visual embedding $\mathbf{z}_v$ with a semantic text embedding $\mathbf{z}_t$ using the pre-trained CLIP model's frozen text encoder. The encoder tokenises a text prompt $t$ before feeding the tokens via a Transformer network. The final text embedding $\mathbf{z}_t \in \mathbb{R}^{D}$ is then obtained by linearly projecting the output embedding corresponding to the special `[EOT]' (End-of-Text) token. Also, to compute the contrastive loss between the multimodal visual embedding $\mathbf{z}_v$ and the text embedding $\mathbf{z}_t$, we perform the L2-normalization represented as $\hat{\mathbf{z}}_v = \frac{\mathbf{z}_v}{\|\mathbf{z}_v\|_2}$ and $\hat{\mathbf{z}}_t = \frac{\mathbf{z}_t}{\|\mathbf{z}_t\|_2}$, respectively. These normalized vectors are used to compute the symmetric contrastive loss, which serves as the optimization objective for the learnable parameters $\theta_V$ and the fusion layer.

\subsection{Network Optimization}
Inspired from CLIP \cite{Clip}, we use the symmetric contrastive loss function to optimise the parameters of our MMLGNet, where we align the text and fused visual embeddings in the shared latent space. Consider a mini-batch of $n$ image-text pairs, $\{(x_{\mathbf{p}_i}, t_i)\}_{i=1}^n$, where $\mathbf{p}_i$ represents the pair of HSI and LiDAR patches for the $i$-th sample, and $t_i$ is its corresponding text description such as $\texttt{"the hyperspectral patch of [CLS]"}$. 
First, we compute a similarity matrix $\mathbf{S} \in \mathbb{R}^{n \times n}$, where each element $S_{ij} =  \frac{\hat{\mathbf{z}}_{v,i} \cdot \hat{\mathbf{z}}_{t,j}^T}{\tau}$ represents the cosine similarity between the $i$-th visual embedding and the $j$-th text embedding, 
where, $\tau$ is a learnable temperature parameter that scales the logits, helping to control the sharpness of the distribution.
Our goal is to maximize the similarity of the positive pairs while simultaneously minimizing the similarity of the negative pairs by reducing the symmetric cross-entropy loss over the similarity scores, defined in Equations \ref{eq:v2t} and \ref{eq:t2v}. \\
\noindent-\textbf{Visual-to-Text Loss ($\mathcal{L}_{v\rightarrow t}$)} computes similarity and generate logits for a classification problem, where the goal is to predict the correct text for each visual input.
\begin{equation}\label{eq:v2t}
    \mathcal{L}_{v\rightarrow t}= -\frac{1}{n} \sum_{i=1}^{n} \log \frac{\exp(S_{ii})}{\sum_{j=1}^{n} \exp(S_{ij})}
\end{equation}
\noindent-\textbf{Text-to-Visual Loss ($\mathcal{L}_{t\rightarrow v}$)} predicts the appropriate visual input for every text prompt by treating each column of the similarity matrix (or each row of $\mathbf{S}^T$) as a logit.
\begin{equation}\label{eq:t2v}
    \mathcal{L}_{t\rightarrow v} = -\frac{1}{n} \sum_{j=1}^{n} \log \frac{\exp(S_{jj})}{\sum_{i=1}^{n} \exp(S_{ij})}
\end{equation}

The final optimization objective is the average of these two losses $\mathcal{L}_{\text{total}} = \frac{\mathcal{L}_{v\rightarrow t} + \mathcal{L}_{t\rightarrow v}}{2}$, ensuring a robust, bidirectional alignment between the modalities, defined in Equation \ref{eq:final_loss}.
\begin{equation}
    \label{eq:final_loss}
    \psi_\theta^* = \arg\min_{\psi_\theta} \mathcal{L}_{\text{total}}
\end{equation}

\section{Experimental Evaluations}
\noindent{\textbf{Dataset description:}}
We experiment our proposed MMLGNet with two benchmark multimodal remote sensing datasets. 
a) \textbf{MUUFL Gulfport} \cite{muuflscenelabels} dataset is acquired over the University of Southern Mississippi contains a 325 × 220 pixel hyperspectral imaging (HSI) with 64 post-processed spectral bands. It also includes LiDAR elevation data from two rasters and covers 11 urban land-cover classes with 1100 training and 52587 testing samples. b) \textbf{Trento} dataset was acquired over rural areas south of Trento, Italy, which consists of 63 HSI bands \cite{trento}. The LiDAR provides a single elevation raster. The 600 × 166-pixel scene includes six mutually exclusive vegetation classes. For our experiments, we used the standard disjoint training and test sets provided with the dataset, which consist of 600 training and 29,614 testing samples. We extract the patch of size 11x11 for all the experiments.
\begin{table}[ht!]
\centering
\caption{Accuracy analysis on the Trento dataset (in \%).}
\label{tab:trento_comparison}
\scalebox{0.9}{
\begin{tabular}{lccccc}
\toprule
    Classes 
      & SVM 
      & ELM 
      & TB-CNN 
      & FusAtNet
      & MMLGNet \\
       
      & \cite{svm} 
      &  \cite{elm} 
      &  \cite{twobranchcnn} 
      & \cite{fusatnet}
      & \\
    \midrule
    Apples    & 85.49 & 95.81 & 98.07 & 98.99 & \textbf{99.95} \\
    Buildings & 89.76 & 96.97 & 95.21 & 99.31 & \textbf{99.68} \\
    Ground    & 59.56 & 96.66 & 93.32 & 95.87 & \textbf{100.00} \\
    Woods     & 97.42 & 99.39 & 99.93 & \textbf{99.93} & 99.89 \\
    Vineyard  & 93.85 & 82.24 & 98.78 & 99.56 & \textbf{99.81} \\
    Roads     & 89.96 & 86.52 & 89.98 & 91.23 & \textbf{95.74} \\
    \midrule
    OA    & 92.30 & 91.32 & 97.92 & 99.06 & \textbf{99.42} \\
    AA       & 86.01 & 92.93 & 96.19 & 98.48 & \textbf{99.18} \\
    $\kappa$& 0.8971 & 0.9042 & 0.9681 & 0.9879 & \textbf{0.9922} \\
    \bottomrule
\end{tabular}
}
\end{table}
\begin{table}[!t]
\centering
\caption{Performance comparison on the MUUFL Gulfport dataset (in \%).}
\label{tab:muufl_comparison}
\scalebox{0.88}{
\begin{tabular}{l c c c c c}
\toprule
    Classes 
      & SVM 
      & ELM 
      & TB CNN  
      & FusAtNet 
      & MMLGNet \\
      
      & \cite{svm} 
      & \cite{elm} 
      & \cite{twobranchcnn} 
      & \cite{fusatnet}
      & \\
    \midrule
    Trees                  & 95.97 & 94.89 & 97.40 & \textbf{98.10} & 89.97 \\
    Grass Pure             & 62.71 & 62.23 & 76.84 & 71.66 & \textbf{86.67} \\
    Grass Ground Surface   & 83.60 & 83.15 & 84.31 & \textbf{87.65} & 77.07 \\
    Dirt and Sand          & 78.60 & 57.88 & 84.93 & 86.42 & \textbf{98.32} \\
    Road Materials         & 92.72 & 93.33 & 93.41 & \textbf{95.09} & 90.80 \\
    Water                  & 95.10 & 68.32 & 10.78 & 90.73 & \textbf{99.18} \\
    Buildings’ Shadow      & 71.23 & 47.01 & 63.34 & {74.27} & \textbf{93.53} \\
    Buildings              & 87.96 & 77.58 & 96.20 & {97.55} & 94.20 \\
    Sidewalk               & 41.11 & 32.15 & 54.30 & {60.44} & \textbf{76.58} \\
    Yellow Curb            & 11.05 &  0.00 &  2.21 & { 9.39} & \textbf{93.98} \\
    Cloth Panels           & 88.76 & 78.29 & 87.21 & {93.02} & \textbf{99.41} \\
    \midrule
    OA
                           & 86.90 & 83.10 & 89.38 & \textbf{91.48} & 88.79 \\
    AA 
                           & 83.37 & 63.17 & 68.26 & {78.58} & \textbf{90.87} \\
    $\kappa$      & 0.8255 & 0.7742 & 0.8583 & \textbf{0.8865} & 0.8542 \\
    \bottomrule
\end{tabular}
}
\end{table}
\noindent\textbf{Architecture details:}
The proposed MMLGNet employs two modality-specific vision encoders for HSI and LiDAR data, each processing $11 \times 11$ input patches and projecting them into a shared 512-dimensional embedding space. The HSI encoder takes an $11\times11\times B$ patch (where $B=63$ for Trento and $B=64$ for MUUFL) and passes it through three convolutional stages. Firstly, we use 64 filters of size $3\times3$, followed by batch normalization, ReLU activation, and $2\times2$ max-pooling, reducing the spatial size from $11\times11$ to $5\times5$. Thereafter we apply 128 filters and further reduces it to $2\times2$, while the final block uses 256 filters to reach $1\times1$. The output is flattened and passed through a fully connected layer, resulting in an HSI embedding $\mathbf{z}^{\mathrm{HSI}} \in \mathbb{R}^{256}$.

Similarly, the LiDAR encoder processes an $11\times11\times1$ DSM patch through three convolutional stages: 32, 64, and 128 filters, respectively, each followed by $2\times2$ max-pooling to obtain a $1\times1$ feature map. This is flattened and passed through a fully connected layer to produce a 256-dimensional LiDAR embedding $\mathbf{z}^{\mathrm{LiDAR}} \in \mathbb{R}^{256}$. These two modality-specific embeddings are concatenated and passed through a linear layer to form the final visual embedding $\mathbf{z}_v \in \mathbb{R}^{512}$. This embedding is L2-normalized and optimized using contrastive loss against the corresponding text embedding $\mathbf{z}_t \in \mathbb{R}^{512}$, obtained from a frozen pretrained CLIP text encoder $f_t$~\cite{Clip}.

\noindent\textbf{Training and evaluation protocols:} We perform all the experiments using on T4 GPU provided by Google Colab(free tier) with Pytorch deep learning framework. The proposed MMLGNet has been optimized using the Adam optimizer \cite{Adam} with a learning rate of \num{1e-4} for maximum of 100 epochs with a batch size of 128. Also, we employed an early stopping mechanism with a patience of 15 epochs to avoid overfitting. The best-performing model has been evaluated on the test set using three standard metrics: Overall Accuracy (OA), Average Accuracy (AA), and the Cohen's Kappa coefficient (k).

\subsection{Comparison to Literature}
We compare the proposed MMLGNet against outperforms several established methods that rely solely on visual modalities using the Trento and MUUFL Gulfport datasets. It is important to note that the primary goal of our proposed MMLGNet is to showcase the advantage of leveraging multimodal information by integrating remote sensing visual data with language modality. Specifically, on the Trento dataset, MMLGNet surpasses SVM \cite{svm}, ELM \cite{elm}, Two-branch(TB-CNN) \cite{twobranchcnn}, and FusAtNet \cite{fusatnet} as shown in Table~\ref{tab:trento_comparison}. We further evaluate MMLGNet on the MUUFL Gulfport dataset, where it achieves a substantial gain in AA, outperforming referred methods by a margin of at least $12\%$. These results signifies the effectiveness of incorporating language-guided information for improved generalization and semantic alignment in multimodal RS.

\subsection{Ablation Studies}
\noindent\textbf{Ablation with loss functions:} As defined in Equation~\ref{eq:final_loss}, the symmetric contrastive loss jointly optimizes visual-to-text ($\mathcal{L}_{v\rightarrow t}$) and text-to-visual ($\mathcal{L}_{t\rightarrow v}$) alignments. We ablate MMLGNet by training with each loss individually and compare the results with the full bi-directional loss, as shown in Table~\ref{tab:loss_ablation}. The symmetric loss consistently yields the best performance, improving average accuracy by $0.16\%$ on Trento and $0.22\%$ on MUUFL over $\mathcal{L}_{t\rightarrow v}$, and by $0.15\%$ and $0.47\%$ over $\mathcal{L}_{v\rightarrow t}$ on Trento and MUUFL, respectively.
\begin{table}[!t]
\centering
\caption{Ablation study on the loss function components.}
\label{tab:loss_ablation}
\scalebox{0.9}{
\begin{tabular}{@{}l c c c|c c c@{}}
\toprule
&\multicolumn{3}{c}{Trento}&\multicolumn{3}{c}{MUUFL}\\\hline
\textbf{Loss} & \textbf{OA} & \textbf{AA} & \textbf{$\kappa$}& \textbf{OA} & \textbf{AA} & \textbf{$\kappa$} \\
\midrule
 $\mathcal{L}_{t\rightarrow v}$ & 99.26 & 98.91 & 0.9901& 88.51 & 90.16 & 0.8499 \\
$\mathcal{L}_{v\rightarrow t}$ & 99.27 & 98.98 & 0.9902& 88.32 & 90.31 & 0.8478 \\
$\mathcal{L}_{v\rightarrow t}+\mathcal{L}_{t\rightarrow v}$& \textbf{99.42} & \textbf{99.18} & \textbf{0.9922}& \textbf{88.79} & \textbf{90.87} & \textbf{0.8542} \\
\hline
\end{tabular}
}
\end{table}

\noindent\textbf{Comparison with Different Text Encoders:}
In Table~\ref{tab:trento_encoder_ablation}, we ablate with different text encoder backbones on the performance of the proposed MMLGNet. Specifically, we experiment with six pretrained and frozen text encoders: BERT-base\cite{DBLP:journals/corr/abs-1810-04805}, RoBERTa-base\cite{DBLP:journals/corr/abs-1907-11692}, ALBERT-base-v2\cite{DBLP:journals/corr/abs-1909-11942}, CLIP RN50, CLIP ViT-B/16, and CLIP ViT-B/32. Our results show that RoBERTa-base and CLIP ViT-B/32 are more effective in aligning multimodal visual features compared to other text encoders, achieving superior performance of at least $0.06\%$ and $0.5\%$ for Trento and MUUFL Gulfport datasets, respectively.

\begin{table}[!t]
\centering
\caption{Performance comparison with different text encoders.}
\label{tab:trento_encoder_ablation}
\begin{adjustbox}{width=0.9\columnwidth} 
\begin{tabular}{@{}l c c c| c c c@{}}
\hline
&\multicolumn{3}{c|}{Trento}&\multicolumn{3}{c}{MUUFL}\\\hline
\textbf{Text Encoder} & \textbf{OA} & \textbf{AA}& \textbf{$\kappa$}& \textbf{OA} & \textbf{AA} & \textbf{$\kappa$} \\
\midrule
BERT Base & 99.36 & 99.01 & 0.9914 &86.47 & 88.52 & 0.8240 \\
RoBERTA Base & \textbf{99.48} & \textbf{99.19} & \textbf{0.9931}  & 86.37 & 88.96 & 0.8235 \\
ALBERT Base-v2 & 99.35 & 98.98 & 0.9913 & 87.17 & 89.71 & 0.8338 \\
CLIP RN50 & 99.32 & 98.98 & 0.9909 & 88.29 & 90.40 & 0.8475 \\
CLIP ViT-B/16 & 99.31 & 98.90 & 0.9908 & 88.03 & 89.40 & 0.8436 \\
CLIP ViT-B/32 & 99.42 & 99.18 & 0.9922 & \textbf{88.79} & \textbf{90.87} & \textbf{0.8542}\\
\hline
\end{tabular}
\end{adjustbox} 
\end{table}

\noindent\textbf{Ablation for multimodal visual features:}
To quantify and demonstrate the effectiveness of our fusion strategy, we perform ablation on our MMLGNet architecture using three configurations HSI, LiDAR and HSI+LiDAR. Table~\ref{tab:arch_ablation_both} presents the performance comparison, where we observe that the multimodal visual features guided by language modality outperform the individual visual modality settings in AA by $9.42\%$ and $28.65\%$ for Trento and MUUFL datasets, respectively.

\begin{table}[!t]
\centering
\caption{Architecture ablation study on Trento and MUUFL datasets.}
\label{tab:arch_ablation_both}
\begin{adjustbox}{width=0.8\columnwidth} 
\begin{tabular}{@{}l c c c| c c c@{}}
\hline
& \multicolumn{3}{c|}{\textbf{Trento}} & \multicolumn{3}{c}{\textbf{MUUFL}} \\\hline
\textbf{Modality} & \textbf{OA} & \textbf{AA} & \textbf{$\kappa$} & \textbf{OA} & \textbf{AA} & \textbf{$\kappa$} \\
\midrule
HSI
  & 93.58 & 89.76 & 0.9140
  & 56.90 & 61.98 & 0.4832 \\
LiDAR
  & 42.30 & 38.84 & 0.3066
  & 14.84 & 25.99 & 0.0904 \\
HSI+LiDAR
  & \textbf{99.42} & \textbf{99.18} & \textbf{0.9922}
  & \textbf{88.79} & \textbf{90.87} & \textbf{0.8542} \\
\hline
\end{tabular}
\end{adjustbox} 
\end{table}

\begin{figure}[t]
  \centering
  \includegraphics[width=0.75\linewidth]{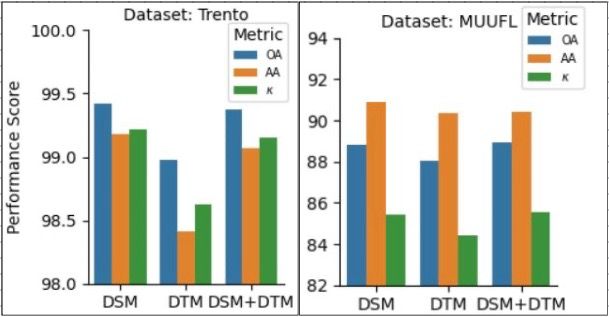}
  \caption{Ablation study on the LiDAR input channels.}
  \label{fig:table6}
\end{figure}

\noindent\textbf{Impact of LiDAR channels:}
We analyze the sensitivity of MMLGNet for LiDAR modality by varying the number of LiDAR input channels i.e., digital surface model (DSM) and digital terrain model (DTM). We experimented using DSM, DTM and DSM + DTM. We found that DTM + DSM setting slightly outperforms OA by $0.05\%$ in Trento dataset, whereas for MUUFL dataset, DTM + DSM performs better than DSM on OA by $0.17\%$, shown in Figure ~\ref{fig:table6}.

\begin{figure}[ht]
  \centering
  \includegraphics[width=\linewidth]{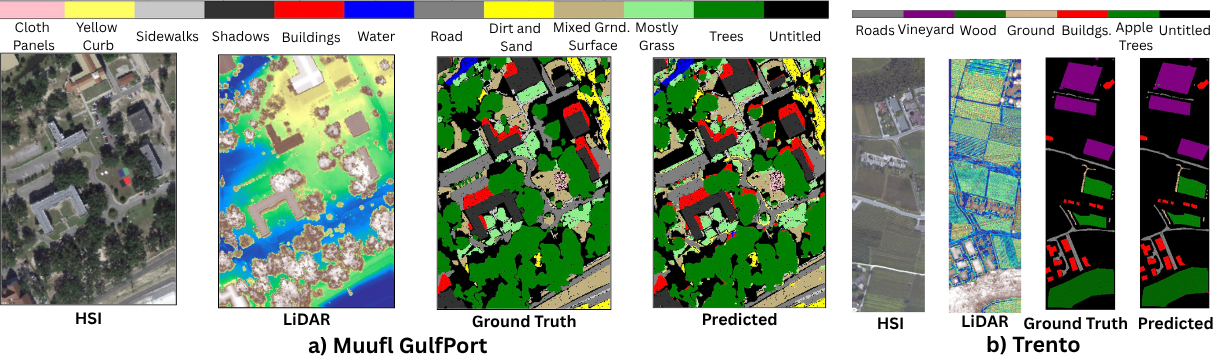}
  \vspace{-0.5cm}
\caption{Classification maps for HSI, LiDAR, ground truth, and MMLGNet predictions on (a) MUUFL and (b) Trento datasets.}
  \label{fig:maps}
\end{figure}

\noindent\textbf{Output classification maps:} In Figure \ref{fig:maps}, we showcase the output classification maps generated from our proposed MMLGNet in the a) Trento and b) MUUFL datasets, which clearly demonstrates the class mappings learned under the language-guided multimodal setup.

\section{Conclusion and Future Work}
The primary goal of this work is to demonstrate the value of integrating textual supervision into multimodal remote sensing (RS) using a simple CNN-based visual architecture. To achieve this, we proposed MMLGNet, a simple framework that aligns visual features from HSI and LiDAR with textual embeddings through contrastive learning. Inspired by CLIP, our method optimizes modality-specific visual features against handcrafted text prompts, enabling semantically rich and unified representations. We evaluate MMLGNet on two benchmark multimodal RS datasets, where it consistently outperforms several methods and improves interpretability. Through modality-aware encoders and contrastive alignment, MMLGNet effectively bridges the semantic gap across modalities. Our ablation studies further validate the importance of textual guidance in enhancing performance. In future, our goal is to go beyond manually made prompts and look into prompt learning methods, which could improve performance and adaptability even more.

\small
\bibliographystyle{IEEEtranN}
\bibliography{references}

\end{document}